\documentclass[runningheads]{llncs}
\usepackage{makeidx}
\usepackage{graphicx}
\usepackage{amsmath,amssymb} % define this before the line numbering.
\usepackage{color}
\usepackage{amsfonts}

\begin{document}
\pagestyle{headings}
\mainmatter

%
% Insert your submission number here
%
%\def\DAGMSubNumber{***}  

%
% Replace with your title
%
\title{Committees of deep feedforward networks trained with few data}

%
% DO NOT MODIFY these for the draft version that is used for the
% review process.
% 
%
%\titlerunning{GCPR YRF 2014 Submission} % \#\DAGMSubNumber{}. CONFIDENTIAL REVIEW COPY.}
%\authorrunning{GCPR YRF 2014 Submission} % \#\DAGMSubNumber{}. CONFIDENTIAL REVIEW COPY.}
%\author{Bogdan Miclut}
%\institute{Institut f\"ur Neuro- und Bioinformatik, Universit\"at zu L\"ubeck}

\author{Bogdan Miclut$^{a}$, Thomas K\"aster$^{a, b}$, Thomas Martinetz$^{a}$, and Erhardt Barth$^{a}$
} 

\institute{
$^{a}$Institute for Neuro- and Bioinformatics, University of L\"ubeck, Ratzeburger Allee 160, 23562 L\"ubeck, Germany \\
$^{b}$Pattern Recognition Company GmbH, Maria-Goeppert Stra\ss e 23562 L\"ubeck, Germany \\
} 

\maketitle

\begin{abstract}
Deep convolutional neural networks are known to give good results on image classification tasks. In this paper we present a method to improve the classification result by combining multiple such networks in a committee. We adopt the STL-10 dataset which has very few training examples and show that our method can achieve results that are better than the state of the art. The networks are trained layer-wise and no backpropagation is used. We also explore the effects of dataset augmentation by mirroring, rotation, and scaling.
\end{abstract}

\section{Introduction}

Recently, deep convolutional neural networks have been shown to perform very well on various challenging pattern-recognition benchmarks. Such networks trained in a supervised way via backpropagation achieve state of the art performance on Caltech-101 \cite{CalTech101}, Caltech-256 \cite{CalTech101}, PASCAL VOC dataset \cite{Everingham10}, MNIST \cite{lecun-98}, ImageNet \cite{imagenet_cvpr09}. However, the drawback of this approach is the requirement of vast amounts of labeled data that is not always available. 

This paper regards unsupervised training of deep neural networks and investigates whether a voting scheme (by a committee of networks) can improve the classification result. We test our method on the STL-10 dataset \cite{Coates_ananalysis}, because it has only a small number of labeled training data. 

For filter training, we use k-means as in \cite{Coates_ananalysis}. After the convolutional step, we found that the local normalization presented by \cite{jarrett-iccv-09} improves the classification result. For the connection between layers, we adopt the random grouping of \cite{DBLP:journals/corr/CulurcielloJDB13}.

Recent state of the art result of \cite{DBLP:journals/corr/DosovitskiySB13} on STL-10 dataset proves that methods from supervised training can be adapted for unsupervised training of networks. Their method creates virtual classes by largely augmenting single images, then training networks on each of these virtual classes using backpropagation. In our paper, we show that better results can be obtained without using backpropatagion.

Another important result on STL-10 is presented in \cite{EDCUFWM} where filters are trained layer-wise: in the first layer, filters are learned via k-means, while in the second layer, filters are being supervised trained via Fisher weight maps for maximizing the between-class distance of descriptors obtained after the second layer. In our work, we only perform layer-wise unsupervised learning of filters. Finally, we show that a committee of networks improves the classification result.
\vspace{-0.5cm}

\section{Network}

We view the network as a chaining of two stages: a feature extractor and a classifier.
The term unsupervised refers to the first stage, which is blind to image labels. The output of the feature extractor is a set of descriptors (one descriptor for each input image). The descriptors of the training image set are used to train the classifier, which will, in the end, assign a label to a descriptor corresponding to a test image.

The feature extractor consists of one or more almost identical layers.
In the following we will present the operations that are being done by such a layer.
We define a feature map as a 2D array given as output by a layer of the network, when presented with one image as input. Thus, an input image is characterized by a set of feature maps given as output by any of the layers of the network.
The main goal is to make these representations invariant to certain transformations, such as translation, scaling, rotation.

\subsection{Preprocessing and Filter Training}
Let $F=\left \{ f_i|f_i\in \mathbb{R}^{m_k\times n_k\times l_k} \right \}$ be the input set of layer $k$.
Here $m_k, n_k$ represent the 2D dimensions of one feature map, $l_k$ is the number of feature maps in layer $k$ and $i=1, ..., N_{train}$ where $N_{train}$ is the number of training images considered. (in the case of the input of the first layer, $l_1=1$ and $m_1, n_1$ are the dimensions of the images)

From the set $F$, we extract a set of patches $X=\left \{ x_k|x_k\in \mathbb{R}^{p\times p\times l_2} \right \}$; for simplicity we extract volume patches consisting of $p\times p$ squares across $l_2$ feature maps (for the first layer, $l_2=1$, therefore each patch is a standard $p \times p$ square).
The elements of $x_k$ are unrolled, thus forming the matrix $X\in \mathbb{R}^{(p \cdot p \cdot l_2)\times N_{patches}}$, where $N_{patches}$ is the number of patches extracted. Each column of $X$ is a $x_k$.

We employ patch-wise normalization as follows: we scale each patch by dividing by the maximum of the absolute value of its elements, then we center each patch by subtracting its mean.
After this, we do ZCA whitening. 

Filters are trained using k-means clustering on the preprocessed patches. Thus, we obtain $U=\left \{ u_k|u_k\in \mathbb{R}^{(p\cdot p\cdot l_2)\times K} \right \}$, where $K$ is the number of trained filters.

\subsection{Convolutions}
After learning the filters, we do a dense feature extraction: for each patch in the input feature maps, we apply all filters via dot product: $y_{ik}=<x_i,u_k>=\sum_jx_{ij}\cdot u_{kj}$. Up to this point, the network is doing pattern matching.

\subsection{Rectification}
We use absolute value as the simplest nonlinearity function. In addition to simple rectification (taking the absolute value) we use ON-OFF separation, where the values $max(0,x)$ and $max(0,-x)$ are feeded separately into the next stage.

\subsection{Local Contrast Normalization}
This step was adopted from \cite{jarrett-iccv-09}. It performs local subtractive and divisive operations.
Let $x_{ijk}$ be the set of feature maps obtained after the \textit{rectification} stage for one input image.
Then, we have: $v_{ijk}=x_{ijk}-\sum_{ipq}w_{pq}\cdot x_{i,j+p,k+q}$, where $w_{pq}$ is a Gaussian weighting window (of size $S\times S$, $S$ depending on the size of the input) normalized so that $\sum_{ipq}w_{pq}=1$. This subtractive operation is a form of edge detection.
For the divisive normalization, we have: $y_{ijk}=v_{ijk}/max(c,\sigma _{jk})$, where $\sigma _{jk}=(\sum_{ipq}w_{pq}\cdot v_{i,j+p,k+q}^2)^{1/2}$. This divisive operation is similar to automatic gain control.

\subsection{Pooling}
We do pooling only within a feature map (across a 2D domain). Pooling is done within patches of size $p\times p$ with stride $s$ (typically $s=p$, meaning pooling is done on disjoint neighboring patches).

Let $\vec{x}$ be a 2D patch. As a pooling function we implemented: $y=(\sum_i x_i^\alpha)^{1/\alpha}$. Such we can move from average pooling by setting $\alpha=1$ to max pooling, for a large $\alpha$.

The output of the pooling stage is also the output of a layer of the network. We can view a layer of the network as a black box, having a set of feature maps as input, and yielding $K$ (the number of trained filters) smaller feature maps as output.

\subsection{Connection between Layers}
This section addresses the issue of layer interconnectivity. The first layer gives for each input images a set of $K_1$ feature maps, therefore, we will have a total number of $K_1 \times N_{training}$ feature maps. ($K_1$ is the number of filters trained in the first layer and $N_{training}$ is the number of images in the training set)

We implemented the \textit{random grouping} explored by \cite{DBLP:journals/corr/CulurcielloJDB13}: the $K_1$ feature maps are divided into $\frac{K_1}{n_K}$ groups of $n_K$. Each group will be treated separately, namely, each group will give $K_2$ feature maps as output (where $K_2$ is the number of filters trained for each group; we chose the same number of filters for each group for simplicity). So, as an example, after the second layer, we will have $K_2\times \frac{K_1}{n_K}\times N_{training}$ feature maps, given we present all the training data to the input of the network.

\subsection{Classifier}
As a classifier we use a multiclass one-vs-all linear L2 SVM.
When presented with a test image, the classifier will give a set of $C$ values (scores). Here $C$ is the number of classes. The assigned class will be the position of the maximum among these scores.

\section{Dataset Augmentation}
We consider just a few augmentation transformations: left-right mirroring, rotations of $\pm 10^{\circ}$ and scaling by a factor of $1/3$.

\section{Committee of Networks}
Here, we ignore the inner workings of the network and consider it as a black box that, when presented with an input image, has a chance to predict its correct label.
Now, we can ask whether there is a strategy of combining multiple such black boxes in order to get a higher success rate. In order to investigate this, we will have to look at the output of the classifier.

Consider an abstract classifier that, when presented with an input, gives a set of $C$ scores between some arbitrary $C_{min}$ and $C_{max}$. How does one pick the most likely label? A straight forward answer is to pick the label corresponding to the highest score. Now, two cases can occur:
1) one score is very large in comparison to all the others; 2) one score is largest, but has some other scores very close to it.

If it is the case of 2) and we know that the classifier is not always right, then some of the true labels are hidden in the scores close to the maximum one.
Therefore, we must solve the following problem: the predicted label is false, but the true label has a score close to that of the predicted one.

One way of looking at this is to consider the process of assigning labels as a noisy stochastic process. The goal is to filter the noise out. In order to be able to do this, we need multiple realizations of this process for each input image.

\subsection{Building the Committee}

Training a network with the same parameters does not necessarily give the same descriptors as output. Three reasons for this: the clustering process of k-means, the ordering of filters obtained by k-means and the random grouping of feature maps.

In order to get a wider variety, we use the augmented dataset to train multiple networks; we also vary the parameters, such as pooling size and stride, to obtain even more networks.
If we are to combine values from the classifiers, these values have to be comparable. Thus, we disregard the original meaning of the classifier output and scale the set of values to $[0, 1]$ in order to obtain some scores that will be later combined.

Now, for an input test image, we have $N$ sets of scores in $[0, 1]$. Each set of scores corresponds to the output of one network.

The simplest method to combine the scores is to sum them up.
Let $S_i=[s_{i1}, s_{i2}, ..., s_{iC}]$ be one set of scores, where $i=1..N$ and $C$ is the number of classes.

Then, $S_{committee}\equiv S_c =\sum_{j=1}^N S_j$. The decision of the committee is taken in the same way as before: the highest score gives the class label.

\section{Experimental Results}
In this section we will describe the specific parameters chosen for the network architecture and present the results obtained.

All experiments were done on the STL-10 dataset with networks having 2 layers of feature extractors. We use this dataset to prove that good results can be obtained even with few training examples.
The STL-10 dataset is comprised of a large collection of unlabeled data, which we do not use in our experiments, 5000 labeled training images and 8000 test images. There are 10 predefined folds, each fold containing 1000 training images. We do testing in the following way: train on each fold of 1000 examples and test on the full set of 8000 images; we then report the average success rate over the 10 folds and the standard deviation.
One of the reasons for choosing this dataset is the ratio of training vs test images, which is 1 to 8.

Images are first converted to grayscale. 
For each layer, the input goes through this chain: patch-wise preprocessing (as described above) and filter training, convolution, absolute value rectification, local contrast normalization, average pooling.

For the first layer, we worked with 300 filters of size $16 \times 16$. 
Pooling was done only across the spatial domain. Various pooling sizes and strides have been tried.
In our experiments, average pooling ($\alpha =1$) was performing better than other values of $\alpha$.

For the connection between the first layer and the second one, we employed random grouping: feature maps were stacked together in groups of 4. We experimented other sizes of grouping, and the main finding was that it does not matter so much how many feature maps are grouped together, but it does matter that the dictionary trained for them to be overcomplete.
In the second layer, filters were of size $3 \times 3 \times 4$: $3\times 3$ in spatial dimension, across 4 feature maps. Hence, the dimensionality is 36. Typically we trained 70-80 filters for each group. So the choice for grouping of 4 was a practical one, in order to keep the dimensionality low and for k-means to be able to converge rapidly.

Pooling in the second layer (also average pooling) was done only in spatial region and was kept fixed at $3\times 3$ with a stride of 3. Other pooling sizes have been tried, but no significant difference was observed.

Once several networks were trained, we got the classification results of each one and normalized them in order to create the scores. Then, the scores are added up and the position of the maximum gives the class label.

\setlength{\tabcolsep}{4pt}
\begin{table}
\begin{center}
\caption{Classification accuracy on STL-10 dataset with 2 layer network, using absolute value rectification varying the pooling in the first layer}
\begin{tabular}{l|l|l|l|l|l|p{2cm}}
\hline\noalign{\smallskip}
Network & Scaling & Rectification & Pooling, Stride & No augmentation & Mirroring & Mirroring and Rotations\\
\noalign{\smallskip}
\hline
\noalign{\smallskip}
$N_1$ & no & absolute value & $12\times 12$, $12$ & 60.57 & 63.27 & 63.60\\
\hline
$N_2$ & no & absolute value & $12\times 12$, $8$ & 60.61 & 63.41 & 64.59 \\
\hline
$N_3$ & no & absolute value &$9\times 9$, $9$ & 59.89  & 62.90  &  64.27  \\
\hline
$N_4$ & $1/3$ & absolute value & $4\times 4, 4$ & 58.9 & 60.52 & not tested \\
\hline
$N_5$ & no & ON-OFF & $12\times 12, 12$ & 61.1 & 64.73 & not tested \\
\hline

\end{tabular}
\end{center}
\end{table}
\setlength{\tabcolsep}{1.4pt}

Using these base networks, we build the committee: $N_1$ (mirroring + rotations), $N_2$ (mirroring + rotations), $N_3$ (mirroring + rotations), $N_4$ (mirroring), $N_5$ (ON-OFF + mirroring).
With this committee, by summing up the scores, we get 68.0\%. 
In the committee above if we only keep mirroring as dataset augmentation, we get 67.39\%.

\setlength{\tabcolsep}{4pt}
\begin{table}
\begin{center}
\caption{Classification accuracy on STL-10 dataset}
\begin{tabular}{l|l}
\hline\noalign{\smallskip}
Paper & Result \\
\noalign{\smallskip}
\hline
\noalign{\smallskip}
Unsupervised feature learning by augmenting single
images \cite{DBLP:journals/corr/DosovitskiySB13}  & $67.4 \pm 0.6$ \\
Efficient Discriminative Convolution Using
Fisher Weight Map \cite{EDCUFWM}  & $66.0 \pm 0.7$ \\
Unsupervised Feature Learning for RGB-D Based Object Recognition \cite{bo_iser12} & $64.5 \pm 1$ \\
Discriminative Learning of Sum-Product Networks \cite{NIPS2012_4516} & $62.3 \pm 1$ \\
Selecting Receptive Fields in Deep Networks \cite{NIPS2011_4293} & $60.1 \pm 1$ \\
This paper & \bf 68.0 $\pm$ 0.55 \\
\hline
\end{tabular}
\end{center}
\end{table}
\setlength{\tabcolsep}{1.4pt}

\vspace{-1cm}

\section{Conclusions}
In this paper we showed that combining different networks and employing a voting scheme improves the classification result.
The committee can be constructed from any base network by varying its parameters or presenting as input different transformations applied on the training set.
When building the committee, one has to bare in mind that the output of one network is not independent from the output of the others, thus the performance will not always increase with the number of networks, but will eventually saturate or even decrease if the choice of networks is poor (for example, adding a very bad performing network). In our experiments, the committee always performed better than any of the individual networks.
Note that we achieve results that are better than state of the art by using rather simple two-layer networks for feature extraction and linear SVMs for classification.

\bibliographystyle{splncs03}
\bibliography{egbib}

\end{document}